\documentclass{article}

 \usepackage[main, final]{neurips_2026}

\usepackage[utf8]{inputenc} 
\usepackage[T1]{fontenc}    
\usepackage{hyperref}       
\usepackage{url}            
\usepackage{booktabs}       
\usepackage{amsfonts}       
\usepackage{nicefrac}       
\usepackage{microtype}      
\usepackage{xcolor}         
\usepackage{amsmath}
\usepackage{enumitem}
\usepackage{graphicx}
\usepackage{float}

\usepackage{amsmath}
\usepackage{algorithm}
\usepackage{algpseudocode}
\usepackage{amsfonts}

\usepackage{algorithmicx}
\usepackage{algpseudocode}
\usepackage{amsmath}
\usepackage{amssymb}
\usepackage{xcolor}

\title{Learning Multi-Relational Graph Representations for DNA Methylation-Based Biological Age Estimation
}

\author{Qing Qing$^{1}$, Xikun Zhang$^{2}$, Zhongyuan Zhang$^{1}$, Jiarui Liu$^{1}$, Xingtong Yu$^{3}$, Xiaotao Shen$^{4}$,\\
\textbf{Ziqi Xu}$^{2}$, \textbf{Qixin Zhang}$^{4}$, \textbf{Zhe Wang}$^{1}$, \textbf{Renqiang Luo}$^{1}$\\
$^1$Jilin University, $^2$RMIT University, $^3$The Chinese University of Hong Kong, \\
$^4$ Nanyang Technological University
}

\begin{document}

\maketitle

\begin{abstract}
Aging clocks aim to estimate biological age, a measure of physiological state distinct from chronological age, from observable biomarkers, and are widely used for health assessment and disease analysis. 
DNA methylation is a particularly informative biomarker due to its stability and strong association with aging, and recent learning-based approaches have improved predictive performance. 
However, most existing methods treat CpG sites as independent features, overlooking the complex and heterogeneous biological relationships among them.
We propose RelAge-GNN, a multi-relational graph neural network framework for DNA methylation-based age prediction. Our method constructs three complementary graphs capturing co-methylation patterns, genomic co-localization, and gene-level associations among CpG sites. 
Each graph is modeled by an independent GNN branch, and a learnable gating mechanism adaptively fuses the resulting representations.
Experiments on large-scale datasets show that RelAge-GNN achieves competitive accuracy and stronger correlation with chronological age compared to state-of-the-art methods. 
Moreover, the model exhibits improved sensitivity in detecting age acceleration across diverse disease cohorts, highlighting its potential utility for disease characterization. 
Finally, through post hoc interpretability analyses, we quantify the contributions of different relational structures and CpG sites, providing biologically meaningful insights and suggesting potential directions for aging-related research.
Our code is available at:~\url{https://anonymous.4open.science/r/RelAge-GNN-F1E3/}.
\end{abstract}

\section{Introduction}
\par Aging clock aims to estimate biological age using observable biomarkers, and it has become an important tool for assessing health status and disease risk. 
Compared with chronological age, biological age more accurately reflects an individual’s physiological state and functional decline~\cite{ferrucci2025measure}. 
Among different biomarkers, DNA methylation has emerged as a particularly reliable indicator due to its stability and strong association with aging, making it widely used in aging clock studies~\cite{chan2025dna,ying2024causality,zhang2026robust}. 
Early works mainly rely on statistical models to select key CpG sites and build prediction functions~\cite{horvath2013dna}. 
These models are simple, interpretable, and effective on many public datasets. 
Recently, with the growth of data and computing power, AI-based methods have started to support aging clock construction and show promising improvements~\cite{de2022pan,galkin2021deepmage}.

\par Despite the rapid progress, current AI-assisted aging clock methods still face several limitations. 
First, most approaches treat CpG sites as independent features, while ignoring their structured biological relationship dependencies, such as co-methylation and genomic proximity. 
Second, although deep learning models can capture nonlinear patterns, they often lack clear interpretability and may overfit complex data. 
In addition, different biological relations may contribute differently across samples, but existing models rarely adapt to such variations~\cite{peng2021reinforced}.
These issues limit the ability of current methods to fully capture aging mechanisms.

\par Some recent studies have attempted to address these challenges by introducing more advanced models. 
For example, tree-based methods and deep neural networks improve nonlinear fitting ability and achieve better performance in some cases. 
Graph-based methods further try to model relationships between CpG sites by constructing edges based on biological knowledge. 
These approaches show that incorporating structure can improve both prediction and interpretability. 
However, most existing methods still rely on a single graph or use fixed strategies to combine different types of relations~\cite{ahmed2025graphage}. 
This makes it difficult to distinguish the contribution of each relation and limits model flexibility. 
As a result, the current solutions are still not sufficient for fully exploiting complex biological structures.

\par To address these limitations, we propose RelAge-GNN, a multi-relational graph neural network framework for DNA methylation-based age prediction. 
Specifically, we construct three complementary graphs to model relationships among CpG sites, capturing co-methylation patterns, genomic co-localization (i.e., whether two CpG sites are located on the same chromosome), and gene-level associations (i.e., whether two CpG sites are located within the same gene). 
Each graph is processed by an independent GNN branch based on PNAConv~\cite{corso2020principal} to learn relation-specific structural representations.
We further adopt a learnable gating mechanism to adaptively fuse the representations from different branches at the node level, enabling the model to dynamically adjust the contribution of each biological relationship across samples. 
The fused representation is then used for age prediction via a regression head.

\par Experiments on large-scale datasets demonstrate that RelAge-GNN achieves competitive predictive performance. 
More importantly, the model shows improved sensitivity in detecting age acceleration across diverse disease cohorts, highlighting its potential for disease characterization. 
Building on RelAge-GNN, we extend the GNNExplainer framework~\cite{ying2019gnnexplainer} to develop an interpretability module that quantifies the importance of both relational structures (graph branches) and individual CpG sites (nodes).
Furthermore, post hoc interpretability analyses provide insights into the contributions of different biological relationships and CpG sites, offering meaningful directions for future aging research.
Our main contributions are summarized as follows:
\begin{itemize} [leftmargin=0.5cm]
    \item We identify that current AI-based aging clocks often treat CpG sites as independent features, failing to capture complex biological dependencies. 
    Furthermore, existing methods typically rely on a single graph or fixed combination strategies, which limit their ability to distinguish the varying contributions of different biological relationships across samples.
    \item We propose RelAge-GNN, a novel framework that constructs three complementary graphs to explicitly model co-methylation, genomic co-localization, and gene-level associations. 
    By integrating independent GNN branches with a learnable gating mechanism, our model adaptively fuses these structural representations, allowing for dynamic adjustment of relationship weights at the node level.
    \item Extensive experiments demonstrate that RelAge-GNN achieves competitive predictive accuracy and exhibits superior sensitivity in detecting age acceleration across diverse disease cohorts.
\end{itemize}

\section{Related Work}
\subsection{Aging clock}
\par The aging clock aims to estimate biological age from molecular biomarkers, providing a more accurate reflection of the physiological state than chronological age~\cite{horvath2018dna}. 
Among different biomarkers, DNA methylation~\cite{griffin2024time} is widely used due to its stability and strong correlation with age~\cite{meyer2025aging}. 
Early epigenetic clocks mainly rely on linear models, which are interpretable and effective on many datasets~\cite{kriukov2025computagebench}.
However, these methods treat CpG sites as independent features and fail to capture complex biological relationships. 
Recent studies introduce deep learning to improve nonlinear modeling, but most approaches still follow a feature vector paradigm and underutilize relational structures. 
As a result, there is a growing shift toward relation-aware modeling, where interactions among CpG sites are explicitly considered. 
This motivates the use of graph-based methods to better capture structured biological dependencies.
Although GraphAge employs a graph to explore the interactions among CpG sites, it blurs the distinctions among different biological relationships and lacks fine-grained processing of different relationship types.

\subsection{Graph Neural Network}
\par Graph neural networks (GNNs) naturally model structured data through message passing and are well-suited for representing relationships among CpG sites~\cite{xia2026graph}. 
By combining node features with neighborhood information, GNNs can better capture complex aging signals compared to traditional methods~\cite{yin2023train,chen2023d4explainer}. 
However, existing graph-based aging clock models often rely on a single graph or fixed fusion strategies, limiting their ability to model heterogeneous biological relations. 
To address this issue, we adopt a multi-relational graph modeling strategy and use PNAConv to enhance representation power under varying graph structures. 
This design improves the model’s ability to capture heterogeneous patterns and supports more flexible and interpretable aging clock modeling.
A more detailed overview of aging clock methods and GNNs is provided in Appendix~\ref{sec:related work}.

\section{Preliminaries} 
\subsection{Notations}
\par Unless otherwise stated, we denote sets with copperplate uppercase letters such as $\mathcal{A}$, matrices with bold uppercase letters such as $\mathbf{A}$.
We define the graph as $\mathcal{G} = (\mathcal{V}, \mathcal{E}, \mathbf{X}, \mathbf{A})$, where $\mathcal{V}$ and $\mathcal{E}$ denote the sets of nodes and edges, respectively.
Similarly, $\mathbf{X}$ and $\mathbf{A}$ represent the input node features and edge attributes.
For the $i$-th graph, we consider the $i$-th graph $\mathcal{G}_i = (\mathcal{V}, \mathcal{E}_i, \mathbf{X}, \mathbf{A}_i)$, characterized by its specific edge indices $\mathcal{E}_i$ and attributes $\mathbf{A}_i$.
Regarding the network architecture, $\phi_i$ denotes the $i$-th convolutional layer that produces the embedding output $\mathbf{H}_i$.
Finally, $\alpha_i$ represents the self-learned fusion weight assigned to each respective graph.

\subsection{Graph Construction}
In the graph construction stage, each CpG site is treated as a node, and the complex interactions between CpGs are modeled as edges. 
Three types of edges are established based on biological prior knowledge: co‑methylation, intra‑chromosomal location, and intra‑genic co‑location.
Unlike a single‑graph approach, the three types of relations are organized into three separate graphs, each retaining only one edge type. This design preserves the semantic distinction between relations and avoids conflating heterogeneous connections into indistinguishable edges. 
In the co‑methylation graph, edge weights are assigned as continuous correlation values.
For the intra‑chromosomal and intra‑genic graphs, the edges are of Boolean semantics (representable as 0/1). 
Through this design, the model can simultaneously leverage both statistical association information and structural annotation information.
Detailed descriptions of the process procedure are provided in Appendix~\ref{sec:graph construction}.

\subsection{Aging Clock Metric}
\par The age acceleration (AA) value is the most critical indicator for evaluating the model's performance and biological significance. 
In disease-related case studies, calculating age acceleration allows for in-depth interpretable analysis, enabling the model to learn methylation graph structure patterns specifically associated with biological aging. 
For a sample with true age $y$ and predicted age $\hat{y}$, the age acceleration is defined as:
\begin{equation}
    AA = \hat{y} - y.    
\end{equation}

\par To provide a foundational support and ensure that the discrepancy between predictions and labels remains within a reasonable range, the model is trained to minimize the regression error. 
This supervised learning approach uses the true age of each sample as the label, employing Mean Absolute Error (MAE), Mean Squared Error (MSE), and Fitting Regression Coefficient (FRC) as auxiliary metrics to maintain basic predictive consistency. 
Beyond prediction accuracy, we emphasize age acceleration as a key metric, as it reflects deviations from expected aging trajectories and is closely associated with disease risk and progression.

\par To further enhance model interpretability, this paper introduces a graph interpreter post-training to quantify the importance of nodes, specific attributes, and the three individual graphs~\cite{nandan2025graphxai}. 
This allows us to observe which biological relations contribute more significantly to the age prediction process. 
By computing these importance scores, we can identify the key nodes and structural drivers that most influence the model's assessment of aging.

\begin{figure}[t]
    \centering
    \includegraphics[width=0.95\textwidth]{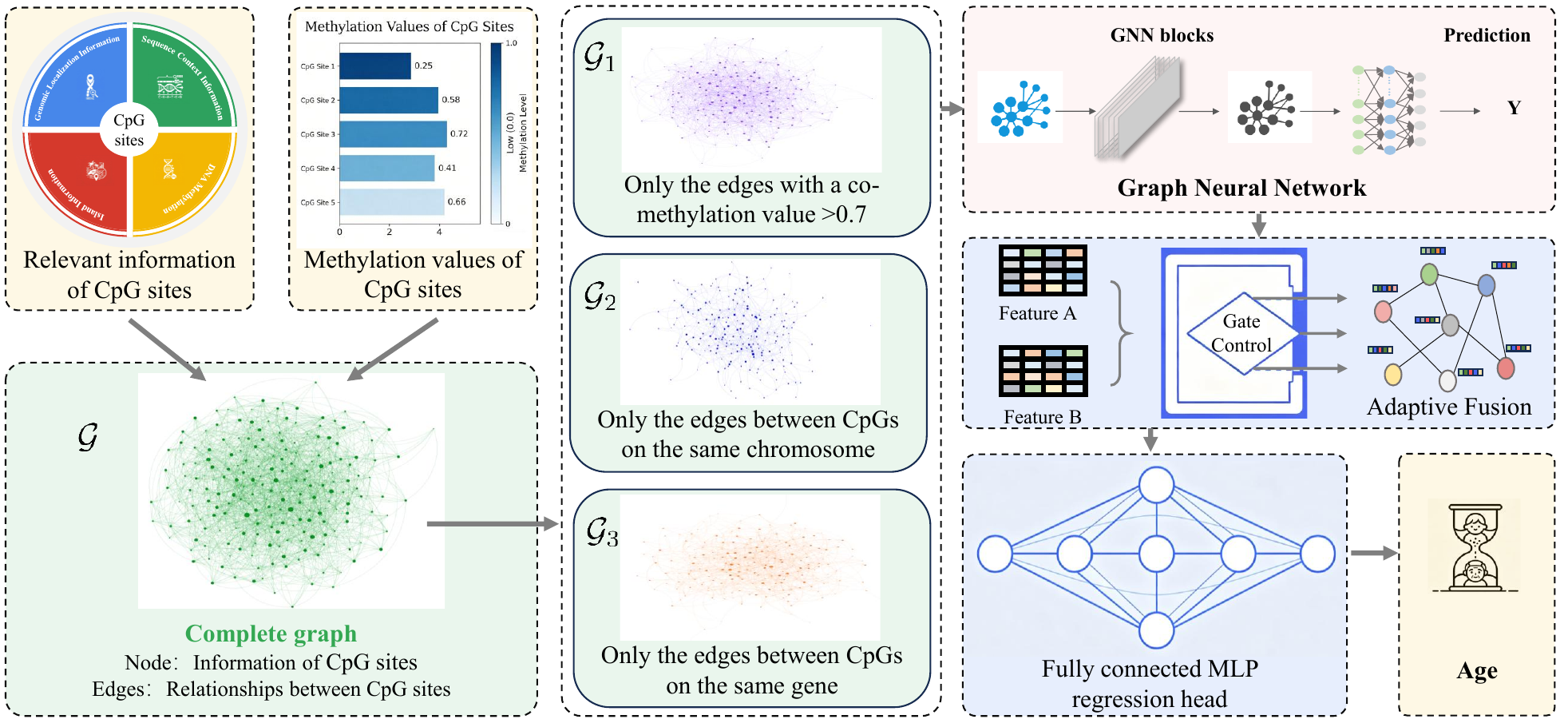}
    \vspace{-1em}
    \caption{The framework of RelAge-GNN.}
    \label{fig:framework}
    \vspace{-1em}
\end{figure}

\section{Methodology}
\par In this section, we present the proposed RelAge-GNN model for age prediction using three-graph methylation data. 
This model is designed for DNA methylation-based age estimation and extends the traditional single-graph modeling approach to jointly model three types of biological relationships: co-methylation, same-chromosome, and same-gene relationships. 
Unlike methods that uniformly treat different edge types, RelAge-GNN designs relatively independent representation learning branches for each edge relationship and integrates multi-relationship information through a learnable fusion mechanism. 
In this way, it preserves the structural semantics of biology while improving prediction accuracy and interpretability. 
The following is an introduction to the network structure of the RelAge-GNN model.
The Corresoongding pseudocode is presented in Appendix~\ref{sec:pseudocode}.

\par The network architecture of the model employs a three-branch parallel encoding framework to simultaneously model different biological relationships among CpG sites.
The co-methylation graph branch focuses on modeling continuous edge attributes, while the same-chromosome and same-gene branches concentrate on learning topological connection patterns.
The node representations output by the three branches are subsequently integrated via a learnable fusion module, rather than using fixed manual weights.
The fused result is then passed through a readout layer and a regression head (MLP) to output the final age prediction, thereby achieving joint modeling of multi-relational information. 

\subsection{Edge-Relation-Specific Three-Branch Encoder}
\par The model employs independent PNAConv layers for each graph. 
While sharing nodes, each graph possesses a unique edge structure.
Let input features be $\mathbf{X}$ with three relational graphs: co-methylation graph $\mathcal{G}_1(\mathcal{V}, \mathcal{E}_1, \mathbf{X},\mathbf{A}_1)$, same-chromosome graph $\mathcal{G}_2(\mathcal{V}, \mathcal{E}_2, \mathbf{X},\mathbf{A}_2)$, and same-gene graph $\mathcal{G}_3(\mathcal{V}, \mathcal{E}_3, \mathbf{X},\mathbf{A}_3)$.
Each convolutional layer first outputs a $64$-dimensional node representation:
\begin{equation}
    \mathbf{H}_1 = \phi_1(\mathcal{E}_1, \mathbf{X}, \mathbf{A}_1), \quad \mathbf{H}_2 = \phi_2(\mathcal{E}_2, \mathbf{X}, \mathbf{A}_2), \quad \mathbf{H}_3 = \phi_3(\mathcal{E}_3, \mathbf{X}, \mathbf{A}_3),    
\end{equation}
where $\phi_k$ denotes the $k$-th PNA convolutional module, and $\mathbf{H}_k$ represents the embedding processed by the $k$-th convolutional layer. 
Each branch is followed by ReLU + Dropout to enhance nonlinear representation and prevent overfitting. 
In implementation, PNAConv employs multiple aggregators (e.g., mean, max, std, min) and multiple scalers (e.g., identity, amplification, attenuation), which improves the expressive power for nodes with different degree distributions, enables more comprehensive information capture, adapts to varying network densities, and achieves greater expressive capability than traditional GNNs.

\subsection{Gated Fusion Module}
\par Unlike fusion with fixed manual weights, this paper employs a learnable gated network to perform node-level adaptive fusion of the three branches~\cite{arevalo2017gated}. 
First, the outputs of the three branches are concatenated as follows:
\begin{equation}
    \mathbf{H}_{cat} = [\mathbf{H}_1 \| \mathbf{H}_2 \| \mathbf{H}_3] \in \mathbb{R}^{N \times 192}.
\end{equation}

\par Then, the gated network is used to obtain the three-way weights for each node:
\begin{equation}
    \alpha = \text{softmax}(\text{MLP}_{gate}(\mathbf{H}_{cat})) \in \mathbb{R}^{N \times 3}.
\end{equation}

\par The final fusion representation is given as:
\begin{equation}
    \mathbf{H}_f = \alpha_1 \odot \mathbf{H}_1 + \alpha_2 \odot \mathbf{H}_2 + \alpha_3 \odot \mathbf{H}_3, \quad \mathbf{H}_f \in \mathbb{R}^{N \times 64},
\end{equation}
where $\alpha_k$ denotes the weight of the $k$-th graph, corresponding to the branch $\mathbf{H}_k$. 
$\odot$ denotes element-wise multiplication, i.e., $\alpha_k$ performs element-wise multiplication with $H_k$ along the feature dimension (64 dimensions).
Such a design enables the model to dynamically select more informative graphs for different samples and different nodes, thereby improving the effectiveness of fusion.

\subsection{Post-Fusion Compression and Regression Head}
\par The fused node representation is first passed through a linear compression layer ($W_m \in \mathbb{R}^{N \times 1}$) followed by ReLU to obtain a one-dimensional node response:
\begin{equation}
    \tilde{\mathbf{H}} = \text{ReLU}(W_m \mathbf{H}_f + b_m), \quad \tilde{\mathbf{H}} \in \mathbb{R}^{N \times 1},
\end{equation}
where $W_m$ is the weight matrix of the post-fusion linear compression layer, which is responsible for mapping the fused 64-dimensional node representation to a one-dimensional response. 
$b_m$ is the bias term of the post-fusion linear compression layer, which together with $W_m$ performs an affine transformation (linear transformation + bias shift), followed by ReLU activation.

\par Subsequently, it is transposed and fed into an MLP regression head, mapping it to the final age prediction. 
The MLP employs LayerNorm, ReLU, and Dropout, replacing BatchNorm to reduce dependence on batch statistics and enhance training stability under small batch sizes. 
This design of the model not only avoids directly mixing continuous relationships with Boolean relationships but also improves the capability of multi-relationship collaborative modeling by replacing fixed weights with a gating mechanism. 
Furthermore, partitioning the relationships into three separate graphs makes subsequent interpretability experiments more convenient and reasonable.

\begin{table}[t]
    \centering
    \footnotesize
    \caption{Summary of the three disease datasets analyzed in this study}
    \label{tab:datasets}
    \begin{tabular}{lrrrrrrrr}
        \toprule
        Dataset & N & \multicolumn{1}{c}{Age Mean $\pm$ SD} & Age Min & Age Max & Male & Female & Healthy & Disease \\
        \midrule
        GSE$19711$   & $802$  & $65.90$ $\pm$ $8.57$  & $49$  & $91$  & $0$    & $802$  & $536$  & $266$  \\
        GSE$41037$   & $1045$ & $36.11$ $\pm$ $14.70$ & $16$  & $88$  & $690$  & $354$  & $719$  & $326$  \\
        GSE$99624$   & $78$   & $68.28$ $\pm$ $9.99$  & $49$  & $87$  & $14$   & $64$   & $46$   & $32$   \\
        \bottomrule
    \end{tabular}
    \vspace{-1em}
\end{table}

\section{Experiments}
\subsection{Datasets}
In our study, we utilized data from the Gene Expression Omnibus (GEO) at the National Center for Biotechnology Information (NCBI) and the ArrayExpress database at the European Bioinformatics Institute (EBI), with a specific focus on blood tissue data~\cite{ahmed2025graphage}.
The final dataset comprises samples from $37$ different datasets, including a total of $3,707$ healthy samples and $624$ disease samples (totaly $4,331$ nodes in the graph).
The healthy samples were used for model training and testing, while the disease samples served primarily as a test set for case studies to examine any unexpected behaviors of the model when predicting on disease samples. 
In this paper, three disease categories are analyzed: $266$ samples from postmenopausal women with ovarian cancer, $326$ samples from patients with schizophrenia, and $32$ samples from patients with osteoporosis.
Table~\ref{tab:datasets} presents the relevant information for the three datasets analyzed in this paper, namely ovarian cancer, schizophrenia, and osteoporosis. 
Information for the remaining datasets can be found in the Appendix~\ref{sec:dataset}.

\begin{figure}[t]
    \centering
    \includegraphics[width=0.95\textwidth]{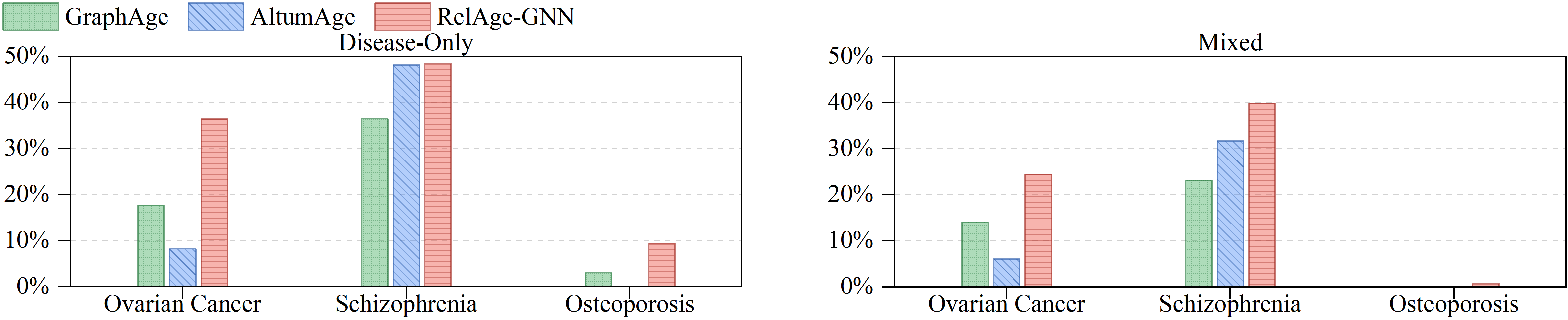}
    \includegraphics[width=0.95\textwidth]{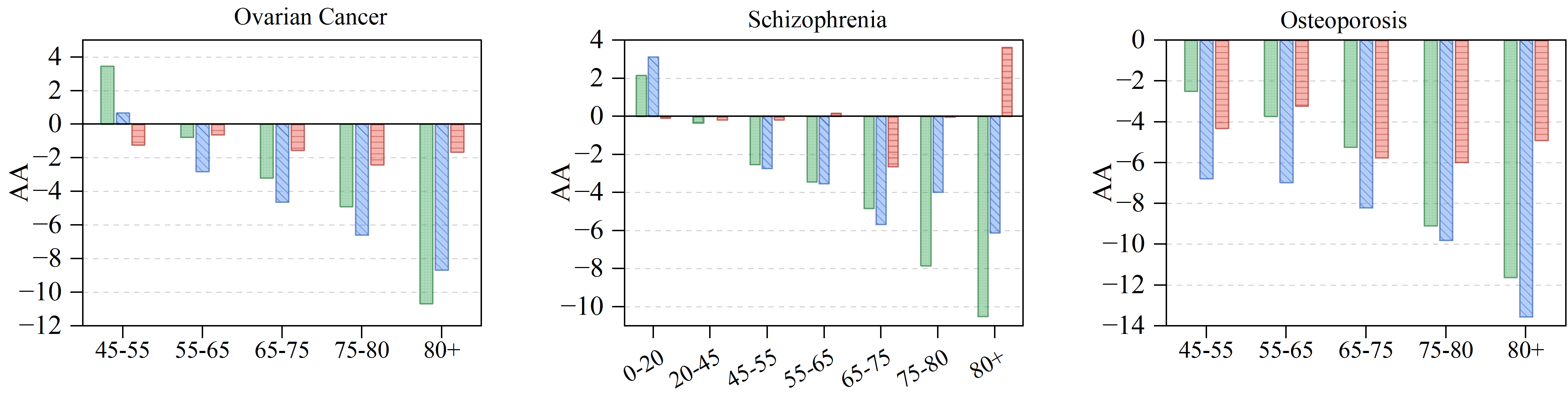}
    \includegraphics[width=0.95\textwidth]{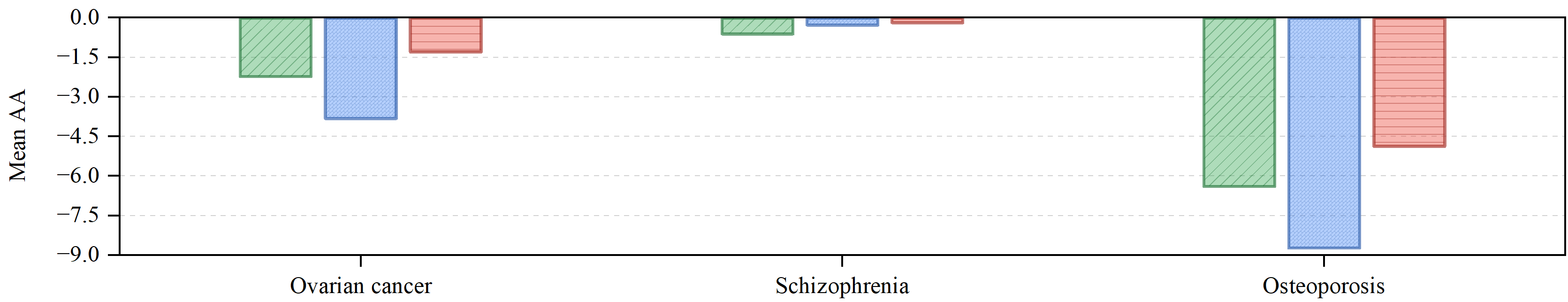}
    \caption{Evaluation of AA as a biomarker and distribution across disease cohorts.
    Top Row: Left: Model sensitivity (proportion of $AA > 0$ in disease samples). Right: Predictive precision (proportion of true disease cases among $AA > 0$ detections).
    Middle Row: Age-stratified AA performance for ovarian cancer, schizophrenia, and osteoporosis.
    Bottom Row: Aggregated performance across all diseases, demonstrating model generalizability in capturing aging patterns.}
    \label{fig:acceleration}
    \vspace{-1em}
\end{figure}

\subsection{Baselines}
\par In our experiments, we compared our model against state‑of‑the‑art baselines in the aging clock field, as well as with the pioneering Horvath clock. 
In total, we selected four representative age prediction baselines covering graph neural network methods, deep learning‑based clock methods, and classical linear epigenetic clock methods: GraphAge, AltumAge, DeepMAge, and Horvath. 
These baselines complement each other in terms of structural modeling capability and model interpretability, providing a relatively comprehensive reflection of performance differences among different technical approaches for DNA methylation‑based age prediction. 
Detailed descriptions of each model are provided in Appendix~\ref{sec:baseline}.

\begin{itemize} [leftmargin=0.5cm]
    \item GraphAge~\cite{ahmed2025graphage} advances methylation age prediction from a vector modeling to a graph modeling paradigm.
    \item AltumAge~\cite{de2022pan} further strengthens the deep learning approach in multi‑tissue scenarios.
    \item DeepMAge~\cite{galkin2021deepmage} is one of the early representative methods that systematically introduced deep learning into DNA methylation age prediction.
    \item The Horvath clock~\cite{horvath2013dna} is one of the foundational works in the field of DNA methylation age prediction.
\end{itemize}

\par We select these four representative baselines to cover three major modeling paradigms: linear elastic net, deep feedforward networks, and graph neural networks.
By training and evaluating them under the same data splitting and preprocessing pipeline, we ensure a fair methodological comparison of model performance.
These baselines not only provide a reference for prediction accuracy but also establish a stable comparative framework for subsequent interpretability analyses, enabling us to better discuss the effectiveness and biological interpretability of our proposed method.
Furthermore, the detailed descriptions of compute resources are provided in Appendix~\ref{sec:compute resources}. 

\subsection{Analysis of Age Acceleration in Disease Cohorts}
\par To evaluate the clinical utility of the proposed RelAge-GNN model, we conducted a comprehensive analysis of AA across multiple disease datasets, including postmenopausal ovarian cancer, schizophrenia, and osteoporosis. 
AA is a critical metric in biological aging research, as it represents the discrepancy between a model's predicted biological age and an individual's chronological age.

\par We first evaluated the models' ability to identify disease-related aging through two distinct experimental setups (Figure~\ref{fig:acceleration}).
In the first experiment, we assessed the proportion of disease samples where the model predicted $AA > 0$. 
For postmenopausal ovarian cancer, RelAge-GNN identifies approximately $40$\% of samples with positive AA, significantly outperforming AltumAge (less than $10$\%) and GraphAge (less than $20$\%).
This finding is consistent with the biological expectation that cancer cell proliferation accelerates systemic aging.
For osteoporosis, RelAge-GNN also shows slightly higher sensitivity than AltumAge and substantially higher sensitivity than GraphAge, while the three models perform comparably on schizophrenia.
The second experiment simulated a realistic diagnostic scenario by mixing disease samples with healthy test sets. 
Among samples predicted to have $AA > 0$, RelAge-GNN achieves a precision of approximately $25$\% for ovarian cancer, which is five times that of AltumAge ($5$\%) and roughly twice that of GraphAge. RelAge-GNN also consistently demonstrates superior accuracy in the schizophrenia and osteoporosis cohorts.

\par A more granular analysis was conducted to compare RelAge-GNN against GraphAge and AltumAge by stratifying the datasets into specific age groups (Figure~\ref{fig:acceleration}).

\begin{itemize} [leftmargin=0.5cm]
    \item Ovarian Cancer: RelAge-GNN generally yielded higher AA values than baselines. 
    In the $55$–$65$ age group, RelAge-GNN predicted an AA of $-0.633$ (the highest among all groups), compared to $-0.783$ for GraphAge and $-2.822$ for AltumAge.
    \item Schizophrenia: In $20$–$45$ age group, all models performed similarly with AA near zero~\cite{wu2021epigenetic}. 
    However, while baselines showed very low AA in older groups, RelAge-GNN maintained stable averages near zero across nearly all age brackets, consistent with the disorder's nature.
    \item Osteoporosis: Although all models produced negative AA values, RelAge-GNN consistently remained closest to the zero baseline, demonstrating its robustness in datasets where substantial age acceleration is not biologically expected.
\end{itemize}

\par Overall, RelAge-GNN achieved the highest mean AA across all disease samples. 
These results indicate that by leveraging triple-graph information, RelAge-GNN captures disease-associated biological aging signals more accurately than existing aging clock models.

\begin{table}[t]
    \centering
    \caption{Comparison on FRC, MAE, and MSE.
    $\uparrow$ denotes the larger, the better; $\downarrow$ denotes the opposite.}
    \footnotesize
    \setlength{\tabcolsep}{12pt}
    \label{tab:my_label}
    \begin{tabular}{lrrrr|r|r}
    \toprule
    Model & Horvath & AltumAge & DeepMAge & GraphAge & Avg & RelAge-GNN \\
    \midrule
    FRC $\uparrow$  & $0.943$ & $0.944$ & $0.952$ & $0.952$ & $0.948$ & $+0.014$ \\
    MAE $\downarrow$ & $3.87$ & $3.30$ & $3.57$ & $3.42$ & $3.54 $ & $-0.11$ \\
    MSE $\downarrow$ & $33.29$ & $28.31$ & $31.14$ & $29.93$ & $30.67$ & $-1.21$ \\
    \bottomrule
    \end{tabular}
    \vspace{-1em}
\end{table}

\begin{figure}[t]
    \centering
    \includegraphics[width=0.95\textwidth]{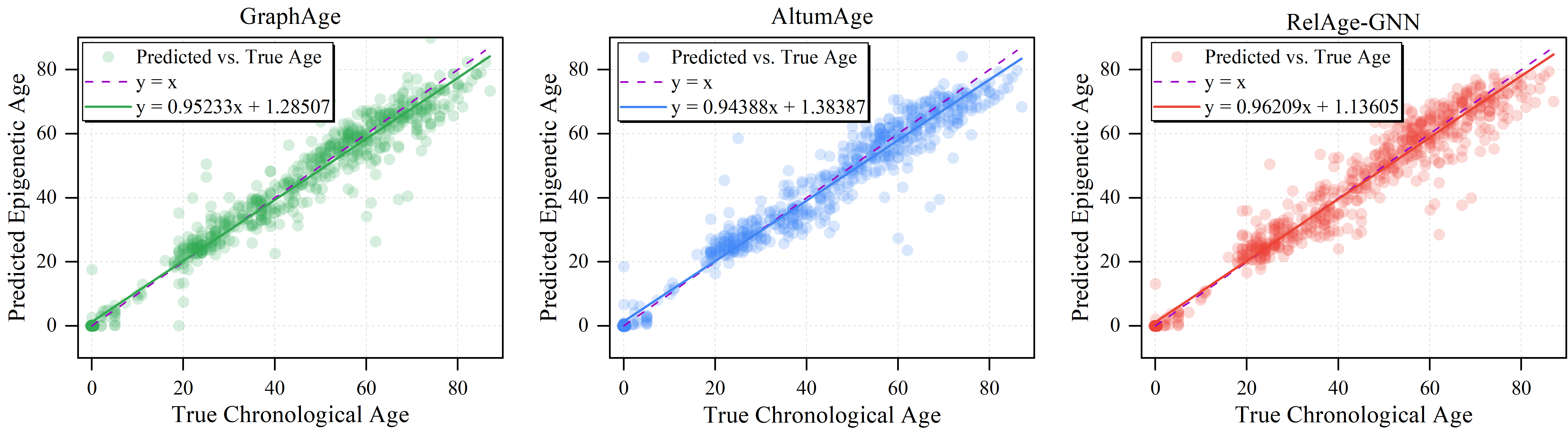}
    \caption{Scatter plots illustrate the correlation between predicted and true age for GraphAge, AltumAge, and the proposed RelAge-GNN. 
    Each panel demonstrates the predictive accuracy of the respective aging clock model across the test cohort.}
    \label{fig:curve fitting}
    \vspace{-1em}
\end{figure}

\subsection{Performance Evaluation on the Test Cohort}
\par To rigorously assess the predictive capabilities of RelAge-GNN, we conducted a comprehensive evaluation using $20\%$ of the total samples as a dedicated test set. 
The remaining $80\%$ of the data followed an $80/20$ split for training and validation, respectively.

\par RelAge-GNN achieves performance comparable to existing baselines across standard statistical metrics, yielding a MAE of $3.4$ and a MSE of $29.46$. 
As illustrated by the scatter plots in Figure~\ref{fig:curve fitting}, the correlation between predicted and chronological ages for RelAge-GNN is remarkably high, with an FRC of $0.962$. 
This represents a quantifiable improvement over established baselines, such as GraphAge ($FRC = 0.952$) and AltumAge ($FRC = 0.944$), underscoring the model's enhanced precision in age estimation.
Crucially, while MAE and MSE reflect basic regression accuracy, they are secondary to $AA$ in capturing the biological significance of epigenetic aging. 
Since age-dependent decline is not inexorable, our model prioritizes the identification of these biological deviations over mere chronological alignment, providing deeper insights into health-related aging patterns~\cite{buckley2023cell}.

\par We further examined the model's robustness across different biological sexes and age intervals.
Within identical age groups, the model exhibits largely consistent performance between sexes, though females show marginally higher MAE and MSE values compared to males.
Predictive accuracy shows a gradual decline as chronological age increases. This trend is consistent with the biological hypothesis that epigenetic noise accumulates over time, leading to greater divergence between chronological age and the epigenetic landscape.

\begin{figure}[t]
    \centering
    \includegraphics[width=0.95\textwidth]{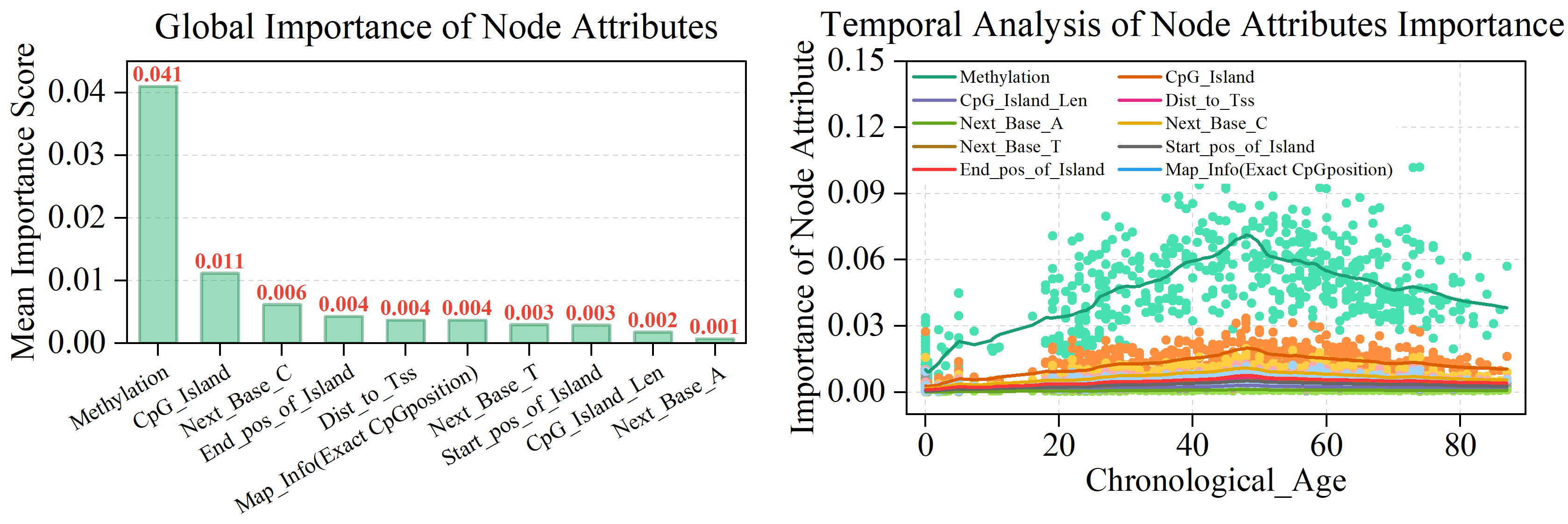}
    \vspace{-1em}
    \caption{Node attribute importance and temporal trends. 
    The left panel displays the importance scores for individual node attributes, while the right panel illustrates their corresponding age-related trends, highlighting how attribute contribution evolves across the lifespan.}
    \label{fig:comparison}
    \vspace{-1em}
\end{figure}

\subsection{Interpretation}
\par We calculated the important nodes and node attributes for each sample to analyze the interpretability of the model at the biological level. 
First, we averaged the importance scores of node attributes across all samples. 
Additionally, we plotted the age-related trends of each node attribute, as also illustrated in figure~\ref{fig:comparison}.
From the average importance scores, it can be observed that the methylation value exhibits the highest importance score, which aligns with our expectations. 
This indicates that methylation carries strong biological significance and contributes importantly to age prediction. 
The Corresoongding pseudocode is presented in Appendix~\ref{sec:pseudocode}.

Furthermore, we found that CpG island-related information (CPG\_ISLAND, end\_pos\_of\_ISLAND, start\_pos\_of\_ISLAND, CPG\_ISLAND\_LEN), next base pair of all CpG sites (Next\_Base\_C, Next\_Base\_T, etc.), and distance to the transcription start site (Dist\_to\_TSS) are also key factors for age prediction.
This reveals a hierarchical pattern of primary signal plus structural context rather than a sole reliance on methylation values. 
Numerous studies have validated the relationship between CpG islands and aging~\cite{christensen2009aging}. 
Variations in CpG island length can affect the complexity of gene regulation and evolutionary mechanisms~\cite{elango2011functional}, thereby influencing the aging process of an individual~\cite{holzscheck2021modeling,perez2020aging}. 
Moreover, since CpG islands are generally located near gene promoters, the starting position of a CpG island becomes particularly important and contributes significantly to age prediction~\cite{illingworth2009cpg}. 
In addition, because the base pair adjacent to a CpG site can influence the methylation status of that CpG site attributes such as Next\_Base\_C and Next\_Base\_T are key nodal factors~\cite{santoni2021impact}.

\par Examining scatter plots of each attribute as a function of age, the importance of methylation attributes rises during the young-to-middle adult years, exhibiting an arch-shaped trend, followed by a slight decline in older age groups while still remaining at high level.
This indicates that the model's reliance on methylation evidence is not constant but rather varies dynamically across different life stages. 
The importance curves for CpG island-related attributes follow a similar direction to that of methylation but with smaller magnitudes, suggesting that CpG island information plays more of an auxiliary role and does not dominate in the age prediction model. 
The remaining attributes generally stay close to baselines, though discrete high-value points can still be observed at specific ages, consistent with the interpretability characteristic of low average contribution but sample-specific local importance.

\begin{table}[t]
    \centering
    \renewcommand{\arraystretch}{1}
    \footnotesize
    \caption{Comparisons among components of the RelAge-GNN.}
    \label{tab:ablation}
    \begin{tabular}{lrrrrrrr}
        \toprule
         & w/o $\mathcal{G}_1$ & w/o $\mathcal{G}_2$ & w/o $\mathcal{G}_3$ & only $\mathcal{G}_1$ & only $\mathcal{G}_2$ & only $\mathcal{G}_3$ & RelAge-GNN \\
        \midrule
        FRC & $1.081$ & $1.007$ & $1.055$ & $0.961$ & $1.023$ & $0.960$ & $0.962$ \\
        MAE & $4.11$  & $3.39$  & $4.11$  & $3.35$  & $3.89$  & $3.57$  & $3.43$ \\
        MSE & $45.84$ & $30.45$ & $39.57$ & $30.1$  & $43.68$ & $33.13$ & $29.46$ \\
        \bottomrule
    \end{tabular}
    \vspace{-1em}
\end{table}

\subsection{Ablation Study}
\par Under a fixed training and evaluation pipeline, we systematically assessed the contribution of each relational graph to model behavior by either removing them one at a time or retaining only a single graph. Model performance was measured using the FRC, MAE, and MSE (see Table~\ref{tab:ablation}).

\par RelAge-GNN, which simultaneously uses $\mathcal{G}_1$, $\mathcal{G}_2$, and $\mathcal{G}_3$, achieved the lowest MSE ($29.46$), with an MAE of $3.43$ and an FRC of $0.962$. 
Overall, it performed best in suppressing large errors and maintaining prediction stability, indicating that the combination of the three graphs helps reduce extreme outlier errors.
When $\mathcal{G}_1$ is removed (retaining only $\mathcal{G}_2$ and $\mathcal{G}_3$) or $\mathcal{G}_3$ was removed (retaining only $\mathcal{G}_1$ and $\mathcal{G}_2$), the MAE increased to $4.11$ in both cases, while the MSE rose to $45.84$ and $39.57$, respectively, representing the worst performance among all leave-one-out settings. 
This suggests that $\mathcal{G}_1$ and $\mathcal{G}_3$ are critical for maintaining accuracy, and the absence of either significantly degrades overall fitting. 
In contrast, removing $\mathcal{G}_2$ (retaining $\mathcal{G}_1$ and $\mathcal{G}_3$) caused relatively minor damage (MAE $3.39$, MSE $30.45$). 
Additionally, some leave-one-out configurations exhibited FRC values greater than $1$ (e.g., $1.081$ for w/o $\mathcal{G}_1$), indicating a systematic slope deviation in the predicted-age linear relationship.
The full model's FRC is closer to $1$, which is more favorable for interpreting a scaling behavior without systematic amplification or compression.

\par When $\mathcal{G}_1$ is used alone, it achieves the lowest MAE ($3.35$) among single-graph configurations, second only to the full model, indicating that $\mathcal{G}_1$ alone contributes the most to model performance. 
The configuration using only $\mathcal{G}_2$ yielded the worst MAE/MSE among single-graph settings, suggesting that chromosomal adjacency structure alone is insufficient to support high-precision age regression. 
The performance of using only $\mathcal{G}_3$ is between ($3.57$ / $33.13$), similarly indicating that relying solely on gene-based relationships is also insufficient to independently support high-precision age regression.

\section{Limitation}
\par Despite the strong performance and design of our method, several limitations remain. 
First, our model is mainly developed and evaluated on blood tissue data, where it already demonstrates stable and competitive results. 
However, biological aging patterns may vary across different tissues, and extending the current framework to multi-tissue settings could further improve its general applicability. 
Second, although we explicitly model three important biological relations and show their effectiveness, these relations still represent a subset of possible interactions among CpG sites. 
There may exist more complex or fine-grained relationships that are not yet included, and incorporating additional biological knowledge could further enhance the model.

\section{Conclusion}
\par We study DNA methylation-based aging clock from a relational modeling perspective. 
We point out that most existing methods rely on feature vector representations and do not fully utilize the biological relationships among CpG sites. 
To address this limitation, we propose RelAge-GNN, a multi-relational graph neural network framework that models co-methylation, same-chromosome, and same-gene relationships in a unified way. 
By constructing three graphs and learning their representations through independent branches, our method is able to capture heterogeneous structural patterns. 
The proposed gating mechanism further enables adaptive fusion of different relations, improving model flexibility and robustness. 
Extensive experiments on datasets show that RelAge-GNN achieves competitive performance and higher correlation with chronological age compared with strong baselines. 
In addition, the model provides meaningful interpretability by quantifying the contributions of different graphs and CpG sites. 
Overall, our work demonstrates that explicitly modeling multi-relational structures is an effective direction for improving aging clock models.
\bibliographystyle{unsrt}
\bibliography{references}


\appendix

\section{Related Work} \label{sec:related work}
\subsection{Aging Clock}
\par The core goal of the aging clock is to estimate an individual's biological age using molecular biomarkers, thereby reflecting the true aging state of the organism rather than merely chronological age~\cite{kriukov2025computagebench}. 
DNA methylation~\cite{seale2022making} stands out from biomarkers such as transcriptomic~\cite{sun2025spatial} or proteomic~\cite{argentieri2024proteomic} signals due to its high cross-tissue measurability, technical maturity, and stable age-related correlations. 
Consequently, it has become one of the most widely used data types for aging clock construction.

\par Classical epigenetic clock methods typically employ linear models (e.g., Elastic Net~\cite{horvath2013dna}), which select CpG sites significantly associated with age and construct regression functions. 
These methods offer strong interpretability, stable training, and good performance on small to medium-sized datasets, and have been validated across multiple public cohorts.

\par However, as data scale and task complexity increase, linear models show clear limitations. 
They struggle to capture nonlinear interactions among CpG sites and treat each site as independent, ignoring structured biological relationships such as co-methylation, same-chromosome, and same-gene associations. 
To address these issues, recent studies have introduced machine learning methods~\cite{dingmachine,sayed2021inflammatory,schultz2020age} such as random forests and deep neural networks, which improve nonlinear modeling capacity~\cite{li2022novel}.

\par Despite these advances, most approaches still rely on feature vector representations and do not explicitly model relationships among CpG sites. 
As a result, aging clock research is gradually shifting from single-site modeling toward relation-aware approaches. 
In this context, graph structures provide a natural way to represent dependencies among CpG sites. 
In this work, we construct a graph-based framework to model these interactions and explore their role in improving age prediction.

\subsection{Graph Neural Networks}
\par GNNs learn representations of structured data through a node–edge–message passing mechanism, making them well-suited for modeling relationships among CpG sites~\cite{luo2026utility}. 
Previous models such as GCN~\cite{kipf2017semi}, GraphSAGE~\cite{hamilton2017inductive}, and GAT~\cite{velivckovic2018graph} have shown clear advantages over traditional vector-based approaches by integrating node features with neighborhood information. 
In aging clock tasks, this allows the model to jointly capture site attributes and their interactions, improving representation power.

\par However, existing graph-based aging clock methods often rely on a single graph structure or simple fusion strategies for multiple relations. 
This makes it difficult to distinguish the contributions of different biological relationships and limits adaptability across samples. 
In addition, conventional GNNs typically use simple aggregation functions such as mean or attention-based weighting, which may not be expressive enough for heterogeneous graphs with varying degree distributions.

\par To address these challenges, we draw inspiration from multi-relational graph learning and decompose methylation relationships into multiple graphs. 
Each graph captures a specific type of biological relation and is processed independently. 
We adopt PNAConv (Principal Neighbourhood Aggregation) as the convolutional operator, which uses multiple aggregators (mean, max, min, std) and scalers (amplification, attenuation) to capture richer neighborhood information. 
This design allows the model to represent central tendency, extreme values, and dispersion while reducing bias caused by degree differences.

\par Compared with traditional GNNs, PNAConv is more robust to heterogeneous graph structures. 
In our framework, this is particularly important because different graphs have different densities and topological properties. 
Furthermore, the multi-branch design preserves the independent contribution of each relation type, leading to better interpretability and more biologically meaningful modeling.

\section{Graph Construction} \label{sec:graph construction}
\subsection{Node Definition}
\par Each CpG site is represented as a graph node $v_i$.
The node features mainly include the following: 
\begin{itemize} [leftmargin=0.5cm]
\item Methylation: The methylation value of each CpG site, which serves as the core epigenetic signal. 
\item Boolean value for being inside a CpG island (CPG\_ISLAND): A Boolean value indicating whether the CpG site is located within a CpG island.
\item Length of the CpG island (CPG\_ISLAND\_LEN): The length of the corresponding CpG island. Variations in island length may affect the complexity of gene regulation. 
\item One-hot encoding of the next base pair: A one-hot encoding of the base pair immediately following the CpG site, as the flanking sequence (i.e., the sequence adjacent to the CpG site) influences its methylation susceptibility. 
\item Starting base pair position of the island (start\_pos\_of\_ISLAND): If the site lies within a CpG island, this attribute denotes the starting base pair position of that island; otherwise, it is set to 0.
\item Ending base pair position of the island (end\_pos\_of\_ISLAND): If the site lies within a CpG island, this attribute denotes the ending base pair position of that island; otherwise, it is set to 0. 
\item Normalized distance from transcription start site (TSS): The normalized distance from the CpG site to the transcription start site of the corresponding gene. If this information is missing, the value is assigned as 1 to indicate the maximum possible distance. As CpG islands are often located in promoter regions, this positional information is particularly important. 
\item Map\_Info: The physical chromosomal position (in base pairs) of the CpG site. This value is normalized and incorporated into the node attributes regardless of whether the site is located within a CpG island.
\end{itemize}

\subsection{Definition of Three Types of Relational Edges}
\par The relationships between two CpG sites are divided into three categories.
The first is the co-methylation value, defined as Comethylation\_value. 
The second is whether they are located on the same chromosome, defined as same\_Chromosome, which is a Boolean value. 
The third is whether they are located within the same gene, defined as same\_Gene, which is also a Boolean value. 
To avoid semantic mixing, we define three graphs based on the above three types of edge relationships: 

\begin{itemize} [leftmargin=0.5cm]
\item Co-methylation graph $\mathcal{G}_1$: This graph retains only those edges where the methylation correlation (i.e., co-methylation value~\cite{affinito2020nucleotide}) between two CpG sites exceeds a pre-specified threshold, with edge attributes being continuous values. 
\item Same-chromosome graph $\mathcal{G}_2$: This graph retains only the edges between two CpG sites located on the same chromosome, i.e., edges where same\_Chromosome = $1$.
\item Same-gene graph $\mathcal{G}_3$: This graph retains only the edges between two CpG sites located within the same gene, i.e., edges where same\_Gene = $1$.
\end{itemize}
Since the co-methylation relationship naturally forms a dense graph, directly connecting all edges would lead to excessive GPU memory consumption and introduce noise.
Therefore, we employ a correlation threshold to sparsify the graph structure. 
Through this construction strategy, the model is able to learn not only the values themselves but also the relationships between values.

\section{The Pseudocode of RelAge-GNN} \label{sec:pseudocode}
\begin{algorithm}
    \caption{RelAge GNN: Age Prediction Process}
    \begin{algorithmic}[1]
        \State \textbf{Input:}
        \State Node features \( \mathbf{X} \in \mathbb{R}^{N \times d_{\text{in}}} \)
        \State Three graphs \( (\mathcal{E}_1, \mathbf{A}_1), (\mathcal{E}_2, \mathbf{A}_2), (\mathcal{E}_3, \mathbf{A}_3) \)
        \State (batch index ignored below for a single graph; training flag controls dropout)
        \State \textbf{Output:}
        \State Scalar graph-level prediction \( \hat{y} \) (flattened 1-D tensor in implementation)
        \State \hspace{0.5cm} \textbf{// Three parallel PNA branches (same $\mathbf{X}$, different topology)}
        \State \( \mathbf{H}_1 \leftarrow \text{Dropout}\big(\text{ReLU}(\text{PNA}_1(\mathcal{E}_1, \mathbf{X}, \mathbf{A}_1)), p\big) \)
        \State \( \mathbf{H}_2 \leftarrow \text{Dropout}\big(\text{ReLU}(\text{PNA}_2(\mathcal{E}_2, \mathbf{X}, \mathbf{A}_2)), p\big) \)
        \State \( \mathbf{H}_3 \leftarrow \text{Dropout}\big(\text{ReLU}(\text{PNA}_3(\mathcal{E}_3, \mathbf{X}, \mathbf{A}_3)), p\big) \)
        \State \hspace{1cm} // \( \mathbf{H}_k \in \mathbb{R}^{N \times 64} \) for \( k = 1, 2, 3 \)
        \State \hspace{0.5cm} \textbf{// Per-node adaptive fusion weights}
        \State \( \mathbf{Z} \leftarrow \text{Concat}(\mathbf{H}_1, \mathbf{H}_2, \mathbf{H}_3, \text{dim}=\text{feature}) \) \hfill // \( \mathbf{Z} \in \mathbb{R}^{N \times 192} \)
        \State \( \alpha \leftarrow \text{Softmax}(\text{Gate}(\mathbf{Z}), \text{dim}=\text{graph}) \) \hfill // \( \alpha \in \mathbb{R}^{N \times 3} \), each row sums to $1$
        \State \hspace{0.5cm} \textbf{// Weighted merge to 64-dim per node}
        \State \( \mathbf{H} \leftarrow \alpha_1 \odot \mathbf{H}_1 + \alpha_2 \odot \mathbf{H}_2 + \alpha_3 \odot \mathbf{H}_3 \) \hfill // \( \mathbf{H} \in \mathbb{R}^{N \times 64} \) (broadcast \( \alpha_k \))
        \State \hspace{0.5cm} \textbf{// Per-node channel reduction}
        \State \( \mathbf{U} \leftarrow \text{ReLU}(\text{MergeLin}(\mathbf{H})) \) \hfill // \( \mathbf{U} \in \mathbb{R}^{N \times 1} \)
        \State \hspace{0.5cm} \textbf{// Graph readout: treat $N$ node scores as one feature vector}
        \State \( v \leftarrow \mathbf{U}^{\mathsf{T}} \) \hfill // \( v \in \mathbb{R}^{1 \times N} \) (row vector of length $N$)
        \State \( \hat{y} \leftarrow \text{MLP}(v) \) \hfill // e.g. Linear(\( N \rightarrow 512 \)) \(\rightarrow\) 1, output scalar
        \State \textbf{return} \(\text{flatten}(\hat{y})\)
    \end{algorithmic}
    \label{alg:relage}
\end{algorithm}

\par RelAge-GNN treats a fixed set of CpG sites as graph nodes. 
The input consists of a node feature matrix \( \mathbf{X} \in \mathbb{R}^{N \times d_{\text{in}}} \) and three relational graphs \( (\mathbf{E}_1, \mathbf{A}_1) \), \( (\mathbf{E}_2, \mathbf{A}_2) \), \( (\mathbf{E}_3, \mathbf{A}_3) \). 
The same set of node features is encoded via three parallel PNA convolutional branches on the three graphs, respectively. 
After passing through ReLU and Dropout, three branch embeddings \( \mathbf{H}_1, \mathbf{H}_2, \mathbf{H}_3 \in \mathbb{R}^{N \times 64} \) are obtained. 
For each node, the three $64$-dimensional vectors are concatenated along the feature dimension to form \( \mathbf{Z} \in \mathbb{R}^{N \times 192} \), which is then passed through a gating network followed by a softmax operation over the graph dimension to produce non-negative mixture weights \( \alpha \in \mathbb{R}^{N \times 3} \) for the three branches on each node, with each row summing to $1$. 
Subsequently, a weighted fusion is performed as \( \mathbf{H} = \alpha_1 \odot \mathbf{H}_1 + \alpha_2 \odot \mathbf{H}_2 + \alpha_3 \odot \mathbf{H}_3 \) to obtain a unified $64$-dimensional node representation. 
Each node is then mapped to a scalar via a linear layer with ReLU, yielding \( \mathbf{U} \in \mathbb{R}^{N \times 1} \). 
Finally, the scalar scores of all nodes are arranged into a vector, transformed into \( v \in \mathbb{R}^{1 \times N} \), and fed into an MLP to output a single graph-level prediction \( y \) (a flattened scalar), which corresponds to the estimated age.
The relevant pseudocode is presented in Algorithm~\ref{alg:relage}.

\begin{algorithm}
    \caption{RelAge-GNN Explainer}
    \begin{algorithmic}[1]
        \State \textbf{Input:} trained model $f_\theta$, graph sample $\mathcal{G}$, integrated graient (IG) steps $S$, branch weights $w_1,w_2,w_3$
        \State \textbf{Output:} node scores $\mathbf{v} \in \mathbb{R}^N$, feature scores $\mathbf{u} \in \mathbb{R}^F$, graph scores $\mathbf{s} \in \mathbb{R}^3$
        \State \hspace{0.5cm} \textbf{// --- A) IGs on node features (joint three graphs) ---}
        \State $x_0 \leftarrow \mathbf{X}$; baseline $\leftarrow 0$ (same shape as $x_0$)
        \State grads $\leftarrow$ empty list
        \For{$k = 0 \ldots S$}
            \State $t \leftarrow k / S$
            \State $x \leftarrow (1-t) \cdot \text{baseline} + t \cdot x_0$; $x.\text{requires\_grad} \leftarrow \text{True}$
            \State $y \leftarrow \text{mean}\big( f_\theta(x, \mathbf{E}_1, \mathbf{A}_1, \mathbf{E}_2, \mathbf{A}_2, \mathbf{E}_3, \mathbf{A}_3) \big)$ \hfill // scalar target, three graphs jointly
            \State $g \leftarrow \partial y / \partial x$
            \State append $g$ to grads
        \EndFor
        \State stack grads $\rightarrow$ $\mathcal{G}_{\text{mat}}$ of shape $(S+1, N, F)$
        \State $\text{avg\_grad} \leftarrow \text{mean over steps of } 0.5 \cdot (\mathcal{G}_{\text{mat}}[k] + \mathcal{G}_{\text{mat}}[k+1])$ \hfill // trapezoidal rule average gradient
        \State $\text{IG} \leftarrow (x_0 - \text{baseline}) \odot \text{avg\_grad}$ \hfill // element-wise, shape $(N,F)$
        \State $\text{abs\_IG} \leftarrow |\text{IG}|$
        \State $\mathbf{v}[i] \leftarrow \text{mean}_f \,\text{abs\_IG}[i, f]$ for $i = 1\ldots N$ \hfill // node importance
        \State $\mathbf{u}[f] \leftarrow \text{mean}_i \,\text{abs\_IG}[i, f]$ for $f = 1\ldots F$ \hfill // attribute (feature) importance
        \State \hspace{0.5cm} \textbf{// --- B) Graph importance via branch occlusion ---}
        \State Define forward decomposition: $h_1 \leftarrow \text{Conv}_1(\mathbf{X}, \mathbf{E}_1)$, $h_2 \leftarrow \text{Conv}_2(\mathbf{X}, \mathbf{E}_2)$, $h_3 \leftarrow \text{Conv}_3(\mathbf{X}, \mathbf{E}_3)$
        \State $y_{\text{full}} \leftarrow \text{mean}\Big( \text{Head}\big( \text{Concat}( w_1 \cdot h_1, w_2 \cdot h_2, w_3 \cdot h_3 ) \big) \Big)$
        \For{$b \in \{1,2,3\}$}
            \State Construct $h$: set $h_b$ to zero, keep other branches unchanged
            \State $y_b \leftarrow \text{mean}\Big( \text{Head}\big( \text{Concat}( w_1 \cdot h_1, w_2 \cdot h_2, w_3 \cdot h_3 ) \big) \Big)$
            \State $\Delta_b \leftarrow |y_{\text{full}} - y_b|$
        \EndFor
        \State $\mathbf{s} \leftarrow \text{normalize}\big( [\Delta_1, \Delta_2, \Delta_3] \big)$ \hfill // e.g., divide by $\sum \Delta + \epsilon$, get relative contribution
        \State \Return $\mathbf{v}, \mathbf{u}, \mathbf{s}$
    \end{algorithmic}
    \label{alg:interpreter}
\end{algorithm}

\par The RelAge-GNN interpreter, applied to a trained model \( f_\theta \) and a single graph sample $\mathcal{G}$, jointly outputs three types of importance: node importance \( v \), node attribute (feature dimension) importance \( u \), and the relative importance \( s \) of the three graph branches. The interpretation consists of two parts.
The relevant pseudocode is presented in Algorithm~\ref{alg:interpreter}.

\par Part 1 (Integrated Gradients):
Under the condition that the structures of the three graphs are fixed, an \( S \)-step linear interpolation path is constructed from a zero baseline to the actual observation \( x_0 = \)$\mathbf{X}$ for the input node feature matrix. 
At each interpolation point \( x \), \( x \) is fed into the forward computations of all three relational graphs simultaneously to obtain a scalar prediction target \( y \) (e.g., taking the mean of the graph-level output), and the gradient \( \partial y / \partial x \) is computed, yielding a sequence of gradients along the path. 
These gradients are averaged along the path using the trapezoidal rule and then multiplied element-wise by \( (x_0 - \text{baseline}) \) to obtain the Integrated Gradients matrix \( IG \), whose absolute value \( |IG| \) is then taken. 
Node importance is obtained by aggregating \( |IG| \) along the feature dimension (e.g., taking the mean), producing a score for each CpG node. 
Feature importance is obtained by aggregating \( |IG| \) along the node dimension, producing a score for each input attribute dimension, thereby distinguishing the contributions of different genomic attributes to the prediction.

\par Part 2 (Graph Branch Masking):
First, the forward pass is decomposed to obtain the representations \( h_1, h_2, h_3 \) from the three convolutional branches. Under the condition that the fusion and regression head remain unchanged, the full prediction \( y_{\text{full}} \) is constructed. 
Then, for each \( b \in \{1, 2, 3\} \), the \( b \)-th branch representation is set to zero while the other branches remain unchanged, yielding the masked prediction \( y_b \). 
The change in prediction after removing the structural information of that graph is measured by \( \Delta_b = |y_{\text{full}} - y_b| \). 
Finally, \( \Delta_1, \Delta_2, \Delta_3 \) are normalized to obtain the relative contribution proportions \( s \) of the three graphs. 
The algorithm ultimately returns \( v \), \( u \), and \( s \), corresponding to node-level, attribute-level, and relational graph-level interpretable evidence, respectively.

\section{Datasets} \label{sec:dataset}
\par The rationale for selecting this dataset is to ensure a fair and credible performance comparison with AltumAge by using the same or similar data sources.
Moreover, compared with models that construct pan-tissue clocks, using only blood tissue data may capture tissue-specific aging-related methylation patterns more effectively than attempting to fit all tissues with a single model.
Blood is one of the most commonly used and easily accessible sample types in biomarker research, and building a model using blood data has high practical application value. 
Therefore, using blood tissue data is more reliable. 
Finally, to construct the graph structure, information on CpG sites common to all samples is required. 
We selected CpG sites shared across all methylation data platforms and obtained structural information such as chromosomal positions and CpG island annotations for these sites from the NCBI GEO supplementary files. 
To compute co-methylation values between CpG sites as edge features, a sufficiently large number of samples is needed to estimate inter-site correlations, and our large-scale healthy sample set makes this feasible.
Table~\ref{tab:datasets_full} presents the relevant information for the $37$ datasets analyzed in this paper. 

\par For missing values, we imputed them using the \( k\)-nearest neighbors algorithm (KNN) with \( k = 5 \). 
Since the original data were derived from two different platforms and were incompatible, we applied the Beta-mixture quantile normalization (BMIQ) method to standardize the data, thereby eliminating discrepancies between platforms and making the data comparable.

\begin{table}[t]
    \centering
    \caption{Summary of the datasets analyzed in this study}
    \footnotesize
    \label{tab:datasets_full}
    \begin{tabular}{lrrrrrrrr}
        \toprule
        Dataset & N & \multicolumn{1}{c}{Age Mean $\pm$ SD} & Age Min & Age Max & Male & Female & Healthy & Disease \\
        \midrule
        E-GEOD-$51388$ & $60$ & $34.52 \pm 12.27$ & $23$ & $74$ & $36$ & $24$ & $60$ & $0$ \\
        E-GEOD-$52588$ & $58$ & $44.40 \pm 18.10$ & $9$ & $83$ & $7$ & $51$ & $58$ & $0$ \\
        E-GEOD-$53128$ & $43$ & $62.22 \pm 6.85$ & $47$ & $76$ & $0$ & $43$ & $43$ & $0$ \\
        E-GEOD-$53740$ & $165$ & $67.88 \pm 10.55$ & $37$ & $93$ & $63$ & $102$ & $165$ & $0$ \\
        E-GEOD-$54399$ & $48$ & $0.00 \pm 0.00$ & $0$ & $0$ & $28$ & $20$ & $48$ & $0$ \\
        E-GEOD-$54690$ & $20$ & $41.30 \pm 8.84$ & $30$ & $57$ & $20$ & $0$ & $20$ & $0$ \\
        E-GEOD-$56553$ & $31$ & $28.89 \pm 9.34$ & $19$ & $47$ & $13$ & $18$ & $31$ & $0$ \\
        E-GEOD-$57484$ & $22$ & $10.58 \pm 0.49$ & $10$ & $12$ & $22$ & $0$ & $22$ & $0$ \\
        E-GEOD-$58045$ & $172$ & $57.24 \pm 8.20$ & $33$ & $80$ & $0$ & $172$ & $172$ & $0$ \\
        E-GEOD-$59509$ & $36$ & $37.89 \pm 13.61$ & $20$ & $59$ & $30$ & $6$ & $36$ & $0$ \\
        E-GEOD-$59592$ & $120$ & $0.38 \pm 0.00$ & $0$ & $0$ & $62$ & $58$ & $120$ & $0$ \\
        E-GEOD-$62219$ & $60$ & $2.22 \pm 1.85$ & $0$ & $5$ & $0$ & $60$ & $60$ & $0$ \\
        E-GEOD-$64495$ & $106$ & $39.40 \pm 17.62$ & $2$ & $74$ & $37$ & $69$ & $106$ & $0$ \\
        E-GEOD-$64940$ & $216$ & $0.00 \pm 0.00$ & $0$ & $0$ & $106$ & $110$ & $216$ & $0$ \\
        E-GEOD-$65638$ & $16$ & $26.25 \pm 4.40$ & $21$ & $32$ & $0$ & $16$ & $16$ & $0$ \\
        E-GEOD-$67444$ & $70$ & $1.22 \pm 1.62$ & $0$ & $5$ & $40$ & $30$ & $70$ & $0$ \\
        E-GEOD-$67705$ & $44$ & $47.73 \pm 11.73$ & $27$ & $66$ & $42$ & $2$ & $44$ & $0$ \\
        E-GEOD-$71245$ & $24$ & $41.92 \pm 8.63$ & $30$ & $52$ & $0$ & $24$ & $24$ & $0$ \\
        E-GEOD-$71955$ & $61$ & $53.56 \pm 11.82$ & $35$ & $79$ & $6$ & $55$ & $61$ & $0$ \\
        E-GEOD-$72338$ & $36$ & $32.25 \pm 11.43$ & $18$ & $52$ & $24$ & $12$ & $36$ & $0$ \\
        E-GEOD-$77445$ & $85$ & $33.80 \pm 15.90$ & $18$ & $69$ & $43$ & $42$ & $85$ & $0$ \\
        E-GEOD-$79056$ & $36$ & $-0.08 \pm 0.10$ & $0$ & $0$ & $17$ & $19$ & $36$ & $0$ \\
        E-GEOD-$83334$ & $30$ & $2.50 \pm 2.54$ & $0$ & $5$ & $12$ & $18$ & $30$ & $0$ \\
        E-MTAB-$2344$ & $24$ & $69.62 \pm 7.08$ & $59$ & $80$ & $12$ & $12$ & $24$ & $0$ \\
        E-MTAB-$2372$ & $47$ & $48.36 \pm 9.90$ & $21$ & $68$ & $24$ & $23$ & $47$ & $0$ \\
        GSE$19711$ & $802$ & $65.90 \pm 8.57$ & $49$ & $91$ & $0$ & $802$ & $536$ & $266$ \\
        GSE$20236$ & $93$ & $62.37 \pm 6.30$ & $49$ & $73$ & $0$ & $93$ & $93$ & $0$ \\
        GSE$20242$ & $40$ & $34.45 \pm 12.65$ & $16$ & $69$ & $11$ & $29$ & $40$ & $0$ \\
        GSE$34257$ & $84$ & $0.00 \pm 0.00$ & $0$ & $0$ & $42$ & $42$ & $84$ & $0$ \\
        GSE$36642$ & $123$ & $-0.07 \pm 0.03$ & $0$ & $0$ & $62$ & $61$ & $123$ & $0$ \\
        GSE$37008$ & $99$ & $32.95 \pm 4.98$ & $24$ & $45$ & $37$ & $62$ & $99$ & $0$ \\
        GSE$41037$ & $1045$ & $36.11 \pm 14.70$ & $16$ & $88$ & $690$ & $355$ & $719$ & $326$ \\
        GSE$49904$ & $71$ & $55.03 \pm 14.53$ & $23$ & $85$ & $22$ & $49$ & $71$ & $0$ \\
        GSE$56606$ & $74$ & $32.32 \pm 10.29$ & $16$ & $52$ & $22$ & $52$ & $74$ & $0$ \\
        GSE$57285$ & $42$ & $43.57 \pm 13.46$ & $19$ & $71$ & $0$ & $42$ & $42$ & $0$ \\
        GSE$69176$ & $150$ & $0.00 \pm 0.00$ & $0$ & $0$ & $89$ & $61$ & $150$ & $0$ \\
        GSE$99624$ & $78$ & $68.28 \pm 9.99$ & $49$ & $87$ & $14$ & $64$ & $46$ & $32$ \\
        \bottomrule
    \end{tabular}
\end{table}

\section{Baseline Model Descriptions} \label{sec:baseline}
\begin{itemize} [leftmargin=0.5cm]
    \item GraphAge advances methylation age prediction from a vector modeling to a graph modeling paradigm: it treats CpG sites as nodes, constructs edges based on co‑methylation, same‑chromosome, and same‑gene relationships, and employs a graph neural network for message passing and age regression. The study shows that graph representations can more directly integrate structural and site‑attribute information, achieving competitive performance and interpretability. However, although GraphAge explores the role of graphs in age prediction, it does not perform fine‑grained processing of different relationships.
    \item AltumAge further strengthens the deep learning approach in multi‑tissue scenarios. The model uses 20,318 cross‑platform CpG sites as input, employs a multi‑layer perceptron (MLP) to learn complex non‑linear mappings related to age, and applies explanation methods such as SHAP to analyze site contributions and interactions among sites. Compared with traditional linear clocks, AltumAge shows significant improvements in multi‑tissue generalization, prediction on elderly samples, and robustness to noise. It is therefore widely regarded as a strong baseline among deep learning‑based methylation clocks. Although AltumAge is representative in terms of non‑linear fitting + interpretable analysis, its relational modeling still relies primarily on implicit feature‑level learning and does not explicitly encode CpG‑CpG topological structures. In this paper, we use AltumAge as one of the core baselines to test whether explicit graph structural modeling can further improve performance and biological interpretability beyond a strong DNN baseline.
    \item DeepMAge is one of the early representative methods that systematically introduced deep learning into DNA methylation age prediction. This work primarily focused on blood samples, using a deep feedforward neural network for regression modeling and adopting a gradient‑based feature selection strategy to select approximately 1,000 key CpG sites from high‑dimensional sites as the final input. It achieved competitive error metrics on independent validation sets and exhibited high age‑acceleration sensitivity in several disease cohorts.
    \item The Horvath clock is one of the foundational works in the field of DNA methylation age prediction. This method was built on multi‑tissue data, using elastic net regression to select 353 clock CpG sites from cross‑platform available sites and predicting age via a weighted linear combination. Its core contribution is the first systematic demonstration that a single model can estimate age across multiple tissues, providing a standard reference for nearly all subsequent methylation clock studies. In terms of methodological attributes, the Horvath model offers strong interpretability and low implementation complexity, but is limited by its linear assumptions, making it difficult to adequately represent high‑order non‑linear interactions among CpG sites. In this paper, we include it as a classical statistical baseline to measure the improvement and practical significance of new methods relative to traditional epigenetic clocks.
\end{itemize}

\section{Compute Resources} \label{sec:compute resources}
\par All experiments were conducted on a high-performance computing cluster to ensure efficiency metrics (training time and memory) are directly comparable.
CPU: Intel(R) Xeon(R) Gold $6455$B.
GPU: NVIDIA L$40$-$48$GB.
Software: Ubuntu $22.04.5$, Python $3.11.14$, PyTorch $2.9.1$+cu$130$, and PyG (PyTorch Geometric) $2.7.0$.

\begin{figure}[t]
    \centering
        \includegraphics[width=0.95\textwidth]{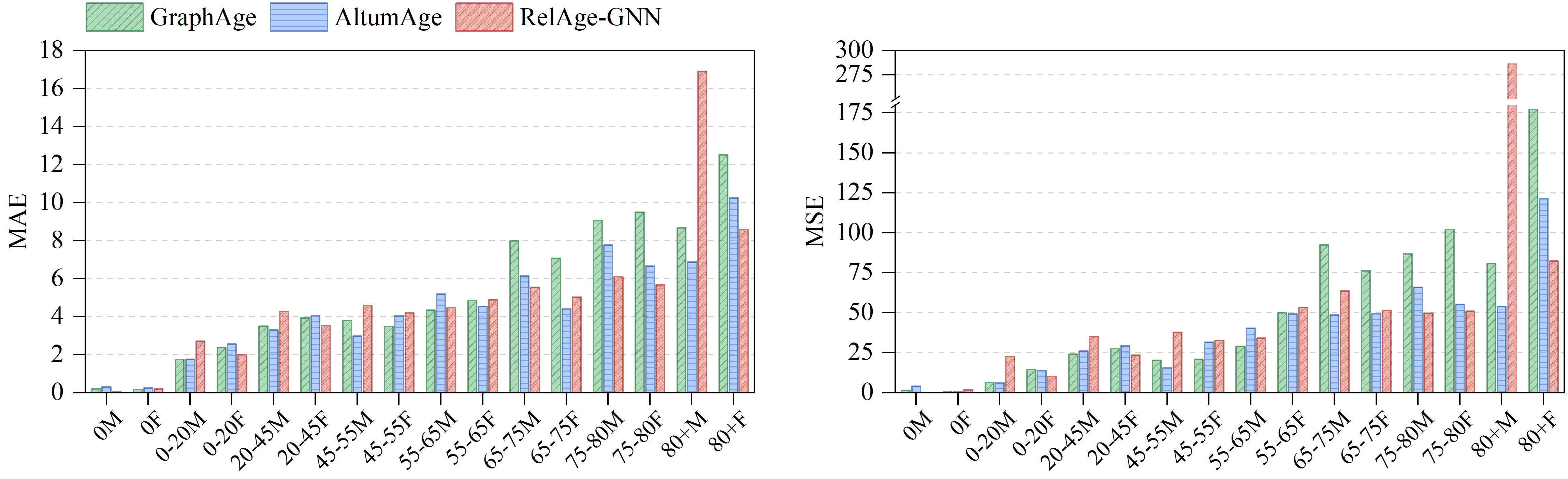}
    \caption{Model performance across age and sex groups. 
    Left: Comparison of MAE across three models stratified by age group and sex. 
    Right: Corresponding MSE results, illustrating the predictive error distribution across demographic subsets.}
    \label{fig:comparison}
\end{figure}

\begin{figure}[t]
    \centering
        \includegraphics[width=0.95\textwidth]{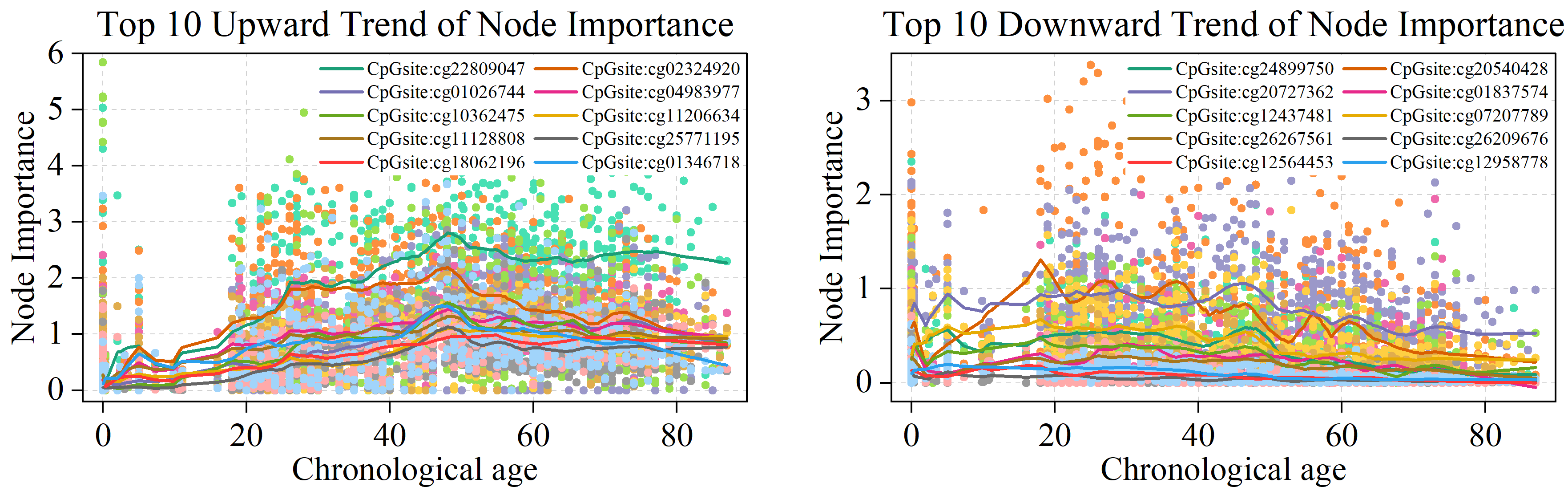}
    \caption{Temporal dynamics of node importance. 
    Left: Age-related trends for the top $10$ nodes with increasing importance across the lifespan. 
    Right: Corresponding trends for the top $10$ nodes with decreasing importance, illustrating the shifting contribution of key features during the aging process.}
    \label{fig:top10_trend}
\end{figure}

\section{Results}
\subsection{MAE and MSE}
We also evaluated the performance of RelAge-GNN, GraphAge, and AltumAge across different age groups and sexes on the test set.
In this experiment, we divided the dataset into different age groups by sex and calculated the mean age acceleration of each model's predictions for each age group.

Overall, all three models exhibited relatively low and comparable MAE values in the young to middle-aged stages (approximately 0–45 years), indicating that within this age range, where samples are relatively abundant, the predicted ages remain within a stable interval. 
As age increases, prediction difficulty generally rises, and MAE and MSE show an upward trend in most stratified groups.
This phenomenon is consistent with the hypothesis that epigenetic noise increases as chronological age deviates from epigenetic age.

Comparing the models, RelAge-GNN tended to maintain lower MAE and MSE values than GraphAge and AltumAge in the middle-aged to elderly stages (approximately $65$–$80$ years).
This demonstrates that our RelAge-GNN model achieves the best performance at higher ages, exhibiting the highest prediction accuracy for elderly populations compared to the other two models.

\begin{figure}[t]
    \centering
        \includegraphics[width=0.95\textwidth]{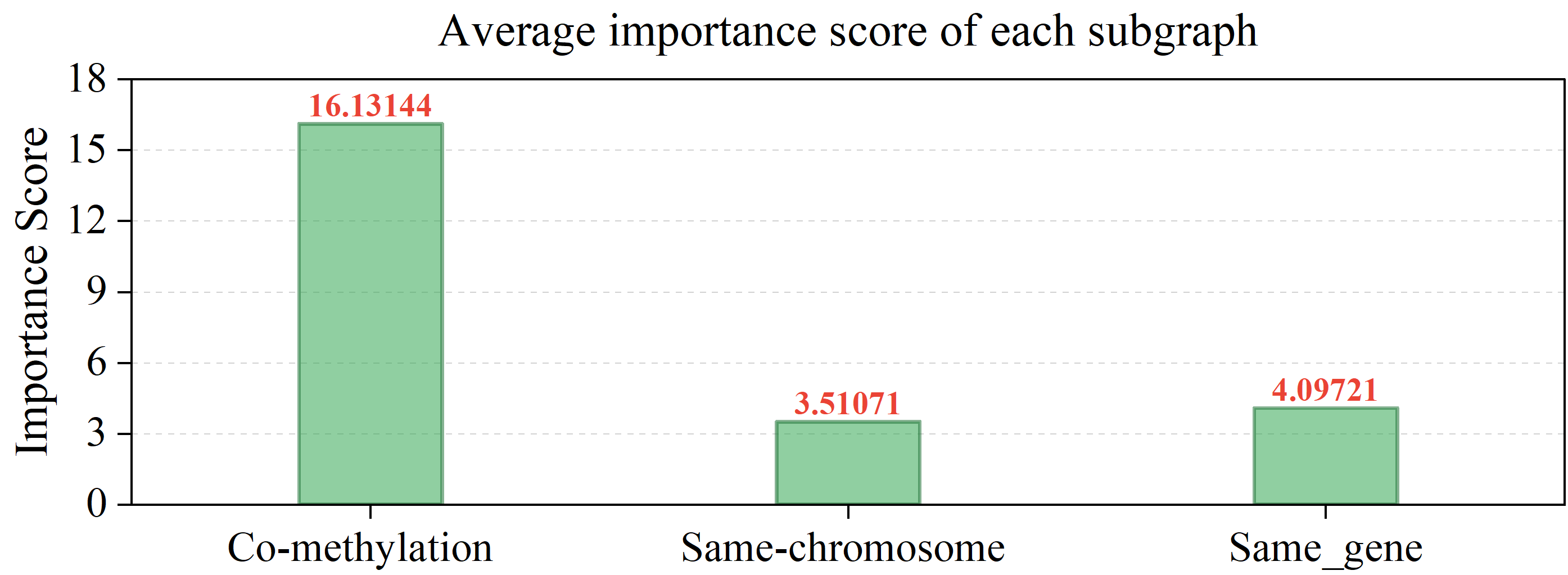}
    \caption{Average importance scores of component graphs. 
    This figure illustrates the mean importance contribution of each of the three graphs within the model architecture.}
    \label{fig:graph_importance}
\end{figure}

\subsection{Node Importance}
In addition to node attributes, we further investigated whether the importance scores of individual CpG nodes exhibit systematic changes with chronological age. 
As shown in the left panel of the figure~\ref{fig:top10_trend}, the importance of ten CpG sites generally increases with age or rises during middle to late life. 
Some fluctuations and peaks are observed from infancy to the preschool years. After entering middle to late adulthood, the trend lines gradually rise, reaching or approaching their peaks around the 40–60 age range. 
In the very elderly stage, some curves show a slight plateau or modest decline, but still remain above the average levels observed in young adulthood. 
Among these, cg$23809047$ maintains the highest importance trajectory across most age groups, indicating that the methylation or neighborhood information carried by this site contributes particularly prominently to the model's discriminative ability in middle to late life.

The right panel of the figure~\ref{fig:top10_trend} shows the opposite trend for ten nodes, where importance tends to be higher during developmental stages and early young adulthood, followed by an overall decline with increasing age. 
In middle to late life, most curves converge to relatively low levels. Some sites (e.g., cg20540428) still exhibit considerable dispersion and local peaks between the ages of 20 and 40, although the long-term trend remains downward. 
Sites such as cg12437481 show a relative peak around the age of 20 and then decline steadily afterward.

\subsection{Graph Importance}
\par Meanwhile, Figure~\ref{fig:graph_importance} presents the average importance scores of the three relational graphs within the interpretability framework. Through this interpretability analysis, we aim to observe which graph contributes the most to the age prediction process. The results show that the importance score of the co-methylation graph is significantly higher than those of the same-gene graph and the same-chromosome graph, indicating that the co-methylation graph plays a dominant role among the three types of relationships. This finding aligns with our expectation that methylation is highly significant for age prediction, suggesting that age-related signals are largely embedded in co-varying methylation networks rather than residing solely in isolated CpG site values.


\end{document}